\documentclass{uai2021} 

\pdfoutput=1

\usepackage[american]{babel}

\usepackage{natbib} 
\usepackage{mathtools} 
\usepackage{booktabs} 
\usepackage{tikz} 
\usepackage{times}
\usepackage{soul}
\usepackage{url}

\usepackage{comment}
\usepackage{graphicx}
\usepackage{subfigure}

\usepackage{hyperref}
\usepackage{float}
\usepackage{stfloats}
\usepackage{booktabs}
\usepackage[ruled,vlined,linesnumbered]{algorithm2e}
\usepackage{verbatim}

\usepackage{amsmath}
\usepackage{amsthm}
\usepackage{comment}
\usepackage{algorithmic}
\usepackage{caption}
\usepackage{amssymb}
\usepackage{makecell}
\usepackage{multirow}
\usepackage{bm}

\newcommand{\ours}{\tt NSRL}

\newcommand{\paul}[1]{{\color{black} #1}}
\newcommand{\paulw}[1]{{\color{black} #1}}

\newcommand{\zhma}[1]{{\color{black} #1}}
\newcommand{\hankz}[1]{{\color{black} #1}}

\title{Learning Symbolic Rules for Interpretable Deep Reinforcement Learning}

\author[1]{Zhihao Ma} 
\author[2]{Yuzheng Zhuang}
\author[3]{Paul Weng}
\author[1]{Hankz Hankui Zhuo}
\author[2]{Dong Li}
\author[2]{Wulong Liu}
\author[2]{Jianye Hao}
\affil[1]{%
    School of Computer Science, Sun Yat-Sen University
}
\affil[2]{%
	Noah’s Ark Lab \\
	Huawei
}
\affil[3]{
	UM-SJTU Joint Institute, Shanghai Jiao Tong University

}

\begin{document}
\maketitle

\begin{abstract}
	Recent \paul{progress} in deep reinforcement learning (DRL) \paul{can be} largely attributed to the use of neural networks. However, this black-box approach fails to explain the learned policy in a human understandable way. To address this challenge and improve the transparency, \hankz{we propose a Neural Symbolic Reinforcement Learning framework by introducing symbolic logic into DRL.} This framework features a fertilization of reasoning and learning modules, enabling end-to-end learning with prior symbolic knowledge. Moreover, interpretability is achieved by extracting the logical rules learned by the reasoning module in a symbolic rule space. \paulw{The e}xperimental results \hankz{\paulw{show} that our framework has better interpretability, along with competing performance \paulw{in} comparison to state-of-the-art} approaches.
\end{abstract}

\section{Introduction}\label{sec:intro}
    Deep reinforcement learning (DRL) has achieved great success in sequential decision-making problems \hankz{such as Atari Games \citep{Mnihet.al2015} and Go \citep{silver2017mastering}}. However, it is hard to \paul{apply DRL to practical problems due notably to its} lack of interpretability.    Interpretability of DRL is important in earning people's trust and developing a robust and responsible system, especially in \hankz{applications related to human safety} such as autonomous driving. Moreover, an interpretable system makes problems traceable and \hankz{debugging easier}. Therefore, interpretability has attracted increasing attention in \paul{the} DRL community recently.
    
    \paul{Interpretability can be either post-hoc or intrinsic, depending on how it is obtained.
    \hankz{For the post-hoc case},
	the black-box model is explained after training by visualizing for instance t-SNE and saliency maps \citep{zahavy_graying_2016} or attention masks 
	\citep{shi_self-supervised_2020}. 
	\hankz{For the intrinsic case,}
	interpretability is entailed by the inherent transparent property of the model \citep{lipton2016mythos}. Our work falls in \hankz{this case}.}
	To improve the \paul{interpretability of DRL}, we investigate \paul{an approach that represents states and actions using first-order logic (FOL) and makes sequential decisions via neural-logic reasoning \hankz{\citep{conference/CIKM/Shi20}}.}
	In this setting, interpretability is enabled by \paul{inspecting} the \paul{FOL} rules \paul{used in the action selection, which can be easily understood and examined by a human.} 
	A number of algorithms \citep{jiang2019neural,Dong.el2019,Payani2003.10386} involving FOL take advantage of neural networks to induce \paul{a} policy \paul{that performs the action selection via approximate reasoning} on symbolic states and \paul{possibly additional} prior knowledge.
    In this context, an action atom with higher confidence of being true is selected after performing some reasoning steps.
    \paul{The rules used in a policy can be learned using a differentiable version of inductive logic programming (ILP) whose goal is to learn FOL rules to explain observed data.
	When a neural network is employed to represent the policy, it can be trained to learn the rules and perform reasoning over those rules by forward chaining implemented in the neural network architecture. The main issues with those approaches are their potential high-memory requirements and their computational costs, which limit their applicability.}
	\paul{Alternatively, \citet{Daoming.el2018} propose a hierarchical reinforcement learning (HRL) approach where a high-level (i.e., task level) policy selects tasks which are then solved by low-level (i.e., action level) policies. The low-level policies interact directly with the environment through potential high-dimensional inputs, while the high-level policy makes decisions via classical planning.
	While this approach can scale to larger problems, 
	it depends on the expert specification of the planning problem to implement the high-level policy.}
	
    To alleviate the \hankz{issues} discussed above, we propose a \paul{novel} framework named Neural Symbolic Reinforcement Learning (\hankz{{\ours}}). In this framework, \paulw{the} policy is induced via a neuro-logic reasoning module without any need of predefined oracle rules or transition model \paulw{specified} in advance, saving expert knowledge dependency compared to \citet{Daoming.el2018}.
    \paul{In contrast to differentiable ILP methods, {\ours} can extract the logical rules selected by the attention modules instead of storing all the rules, thus saving memory budget and improving scalability.
    }
    To the best of our knowledge, this is the first work introducing reasoning into \paulw{reinforcement learning (RL)} that can succeed in complex domains while remaining interpretable. \paul{More specifically, t}his framework features a reasoning module based on neural attention networks, which perform\paul{s} relational reasoning on symbolic states and induce\paul{s} the RL policy. The proposed framework is evaluated on Montezuma's Revenge and Block\paul{s} World. \hankz{The experimental} results demonstrate competing performance with \hankz{comparison to state-of-the-art} RL approaches while providing improved interpretability by extracting the \paul{most relevant relational paths}.

\section{Related Work}\label{gen_inst}

\subsection{Inductive Logic Programming}
	Poor generalization ability and interpretability are common in current  
	machine learning algorithms. Inductive logic programming (ILP), \paulw{an approach} aiming to induce logical rules from data, is promising to address the above mentioned limitations \citep{cropper2020turning}.
	Traditional inductive logic programming approaches require the search in a discrete space of rules and are not robust to noise \citep{Evan.R.el2018}.
	To address those issues, many recent works have proposed various differentiable versions of ILP \citep{Evan.R.el2018,Dong.el2019,Payani2003.10386}. However,
	they are all based on simulating forward chaining and suffer from some form of scalability issues \citep{yuan2019}. In contrast, multi-hop reasoning methods \citep{gardner2015efficient,das2017chains,LaoandCohen.el2010,yuan2019} allow answering queries involving two entities over a knowledge graph (KG) by searching a relational path between them. 
	In the ILP context, such paths can be interpreted as grounded first order rules. 
	Interestingly, they can be computed via matrix multiplication \citep{yang2017differentiable}. 
	Compared to differentiable ILP, multi-hop reasoning methods have demonstrated better scalability. 
	Our work can be seen as the extension of the work by \citet{yuan2019} to the RL setting.

\subsection{Interpretable Reinforcement Learning}
	Recent work on interpretable DRL can be classified into two types of approaches, focusing either on (i) intrinsic interpretability or (ii) post-hoc explanation. 
	Intrinsic interpretability requires the learned model to be self-understandable by nature, which is achieved by using a transparent class of models, 
	whereas post-hoc explanation entails learning a second model to explain an already-trained black-box model. 
	In type (i) approaches, a (more) interpretable policy can be learned directly online by considering a specific class of interpretable policies (e.g., 
	\citep{Daoming.el2018}), 
	or by enforcing interpretability via architectural inductive bias (e.g., 
	\citep{zambaldi2018relational}, 
	\cite{jiang2019neural,Dong.el2019}).
	Alternatively, an interpretable policy can also be obtained from a trained one via imitation learning. 
	\citep{Bastani.el2018, Verma.el2018, Verma_imitation-projected_2019}
	In type (ii) approaches, various techniques have been proposed to explain the policy of DRL agents using t-SNE and/or saliency maps
	\citep{zahavy_graying_2016,greydanus2018visualizing,Gupta.el2020}, 
	attention masks 
	\citep{shi_self-supervised_2020}, 
	visual summaries extracting from histories
	\citep{sequeira_interestingness_2020}, 
	reward decomposition
	\citep{juozapaitis_explainable_2019}, 
	causal model 
	\citep{madumal2020explainable},
	Markov chain 
	\citep{Topin.el2019}. 
	More related to interpretable policies, some work in approach (ii) also tries to obtain a more understandable policy 
	\citep{coppens_distilling_2019} in order to explain a trained RL agent.
	\zhma{Our work falls in the intrinsic case, which preserves interpretability by learning a set of logical rules described by the First-Order Logic.}

\section{Preliminary}
\label{headings}
    In this section, we give a brief introduction to the background knowledge necessary for the proposed framework. Interpretable rules described by First-Order Logic are first introduced, then the basic\paulw{s} of Reinforcement Learning \paulw{(RL)} are briefly \paulw{recall}ed.

\subsection{First Order Logic}\label{FOL}
	A typical First-Order Logic (FOL) system consists of three components: \textbf{Entity, Predicate and \paul{Formula}}. Entities are \paul{constants (e.g., objects)} while a \paul{predicate can be seen as a relation between entities}. An \textbf{atom} $\alpha\paul{=}\paulw{P}(t_1, t_2, .., t_n)$ \paul{is} composed with a $n$-nary predicate $\paulw{P}$ and $n$ terms $\{t_1, t_2, ..., t_n\}$, \paul{where} a term \paul{can} be a constant or variable. An atom is grounded if all terms in this atom are constants. \paul{A \textbf{formula} is an expression formed with atoms, logical connectives, and possibly existential and universal quantifiers. In the context of ILP, one is interested to learn formulas of restricted forms called rules.} 
	A rule \paulw{also} called \textbf{clause} \paulw{can be written as follows:}
	\begin{equation*}
	\alpha \leftarrow \alpha_1 \land \alpha_2, ..., \land \alpha_n
	\end{equation*}
	where $\alpha$ is called \textit{head} atom and $\alpha_1, \alpha_2, ..., \alpha_n$ are called \textit{body} atoms. A clause is grounded with all the associated atoms grounded. The head atom is believed to be true only if all the body atoms are true. For example, $Connected(X,Z) \leftarrow Edge(X,Y) \land Edge(Y,Z)$ is a clause where $X,Y,Z$ are variables and $Connected$, $Edge$ are predicates. If we substitute $X,Y,Z$ with constants $a,b,c$, then $a$ and $c$ \paul{are} considered connected if $Edge(a,b)$ and $Edge(b,c)$ hold. Embedded with prior knowledge, clauses described by FOL are highly understandable and interpretable. \paul{Following \paulw{most} previous neural symbolic approaches, f}unction symbols and recursive definitions \paul{are not} considered in this \paul{work}.

	\begin{figure*}[h] 
		\centering 
		\includegraphics[width=1.0\textwidth]{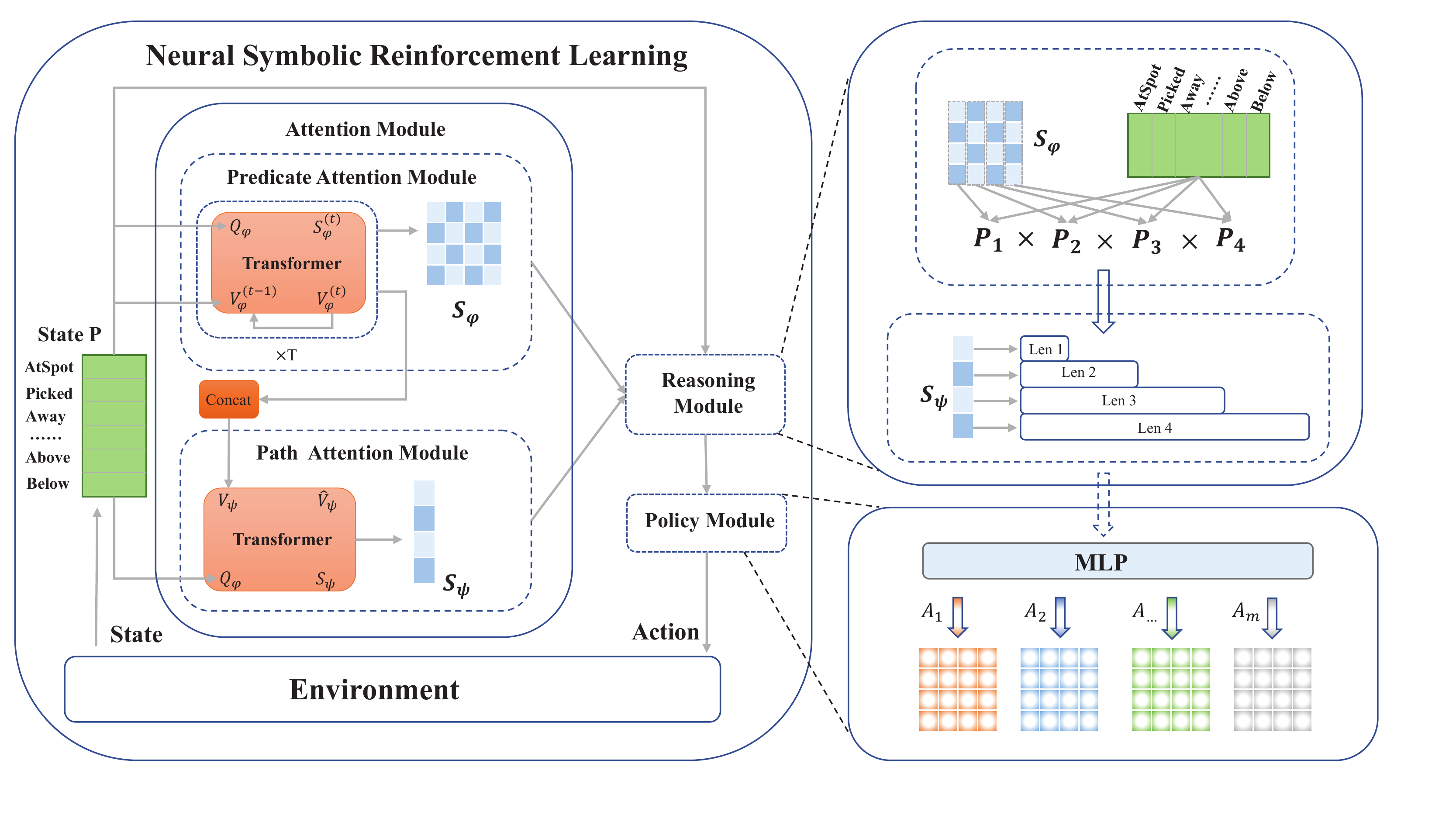}
		\caption{System framework of {\ours}. 
		The left part illustrates \paulw{the} three major components of {\ours}, i.e., reasoning module, attention module, and policy module, which are described in Section~\ref{sec:reasoning module}, Section~\ref{sec:attention module}, \paulw{and} Section~\ref{sec:policy module}, respectively. The right part details the  reasoning module and policy module.}
		\label{Architecture of NSRL} 
	\end{figure*}

\subsection{Reinforcement Learning}
	Consider a Markov Decision Process defined by a tuple $(S, A, P_{ss'}^a, r_s^a, \gamma)$ where $S$ and  $A$ denote the state space and action space, respectively, $P_{ss'}^a$ provides the transition probability of moving from state $s\in S$ to state $s' \in S$ after taking action $a \in A$, $r_s^a$ is the immediate reward obtained after performing action $a$ in state $s$ and $\gamma \in[0,1]$ is a discount factor. The objective of an RL algorithm is to find a \paulw{deterministic} policy $\pi : S \rightarrow A$ \paulw{or a stochastic policy $\pi : S \rightarrow \Delta(A)$ (with $\Delta(A)$ being the set of probability distributions over $A$)} that maximizes the expected return
	$V_{\pi}(s)={\paulw{\mathbb E}_\pi}[\sum_{t=0}^{\infty}\gamma^{t}r_{t} \mid s_0=s]$ where $r_t$ is the reward at time step $t$ received by following $\pi$ from state $s_0\paulw{=s}$. 
	\paulw{The state-action value function is defined as follows: $Q_{\pi}(s, a)={\mathbb E_\pi}[\sum_{t=0}^{\infty}\gamma^{t}r_{t} \mid s_0=s, a_0=a]$.}

\section{Neural Symbolic Reinforcement Learning}
	In this section, we first \paulw{explain} the \paulw{overall} structure of \paulw{our architecture called} Neural Symbolic Reinforcement Learning ({\ours}), \hankz{including the fertilization of three components, i.e., reasoning module, attention module, and policy module. After that, we \paulw{describe} these three components in detail, and present the training process of our {\ours} approach.}

\subsection{System Framework}\label{sec:overall architecture}

	In this section, we describe the structure of {\ours}. 
	As shown in \paulw{Figure}~\ref{Architecture of NSRL}, the symbolic states from \paulw{the} environment \paulw{are} firstly transformed \paulw{into} a matrix \paulw{$\bm P$}, of which each row represents a specific predicate.
	\paulw{This matrix is then} sent to the attention module \paulw{composed of the} predicate and path attention submodule\paulw{s}. 
	The predicate attention submodule iteratively \paulw{processes} matrix $\paulw{\bm P}$ \paulw{resulting in the} generated attention weights $\bm S_\varphi \paulw{= (\bm s_\varphi^{(1)}, \ldots, \bm s_\varphi^{(T)})}$ on predicates at each reasoning step\paulw{, where $T$ is the maximum number of reasoning steps.} Then, the outputs at each step from \paulw{the} predicate attention submodule are \paulw{concatenated} and sent to \paulw{the} path attention submodule to produce attention weights $\bm S_\psi \paulw{= (s_\psi^{(1)}, \ldots, s_\psi^{(T)})}$ on logical rules of different length. \paulw{Next}, matrix $\paulw{\bm P}$ and the attention weights are sent to the reasoning module to perform reasoning on existing symbolic knowledge. 
	As illustrated in the right part \paulw{of Figure~\ref{Architecture of NSRL}}, each column of $\bm S_\varphi$ represents the attention weights on predicates at \paulw{a} reasoning step. Assuming that \paulw{$T=4$}, we denote the predicate matrix at each step as \paulw{$\bm P^{(1)}, \bm P^{(2)}, \bm P^{(3)}, \bm P^{(4)}$}, which are the results of \paulw{the} multiplication of $\bm S_\varphi$ and \paulw{the} symbolic matrix \paulw{$\bm P$}. 
	Then, we sequentially multiply these matri\paulw{ces} to generate logical rules of different lengths.
	\paulw{Next}, we apply path attention weights $\bm S_\psi$ on these rules to generate the reasoning results. 
	\zhma{
	These results are \paulw{then} sent to the multi-layer perceptrons (MLP) in the policy module. 
	Each branch of the \paulw{MLP} output  \paulw{corresponds to an} action predicate.
	\paulw{In the figure, we assume that there are \paulw{in} total $m$ action predicates:} $A_1, A_2, A_{...}, A_m$.
	In the end, we can choose the action atom based on the value of these action predicate \paulw{matrices}.}

\subsection{Reasoning Module}\label{sec:reasoning module}
	Consider a knowledge graph, where objects are represented as nodes and relations are edges. Multi-hop reasoning on such a graph mainly focuses on searching chain-like logical rules of the following form:
	\begin{equation}\label{1}
	\paulw{
	query(x,x') \leftarrow R_1(x, z_1) \land R_2(z_1, z_2) \cdots \land R_n(z_{n-1}, x').}
	\end{equation}
	The task of multi-hop reasoning for a given query corresponds to finding  a relational path from $x$ to $x'$ with multi-steps $x \stackrel{R_1}{\longrightarrow}\cdots\stackrel{R_n}{\longrightarrow}x'$.
	Based on \citet{yang2017differentiable}, the inference of this logical path \paulw{can} be seen as a process of matrix multiplication. 
	Every predicate or relation $P_k$ is represented as a binary matrix $\bm{M}_k$ in $\{0,1\}^{|\mathcal{X}| \times |\mathcal{X}|}$, whose entry $(i,j)$ is $1$ if $P_k(x_i, x_j)$ \paulw{holds, i.e., entity $x_i$ and $x_j$ are connected by edge $P_k$ in the knowledge graph}. 
	\paulw{Set $\mathcal{X}$ contains the objects of the problem}.
	Let $\bm{v}_x$ \paulw{denote} the one-hot encoding of \paulw{an} object $x$. 
	Then, the $t$-th hop of the reasoning along the path can be computed as:
	\begin{align}
	\bm{v}^{(0)} &=\bm{v}_x, \label{2}\\
	\bm{v}^{(t)} &= \bm{M}^{(t)}\bm{v}^{(t-1)}, \label{3}
	\end{align}
	where $\bm{M}^{(t)}$ is the matrix used in $t$-th hop and $\bm{v}^{(t-1)}$ is the path feature vector. 
	After \textit{T} steps reasoning, the score of the query \paulw{for one path} is computed as \paulw{follows}: 
	\begin{equation}\label{4}
	{\rm score(x,x')} = {\bm{v}_{x}}^{\intercal}\prod_{t=1}^T\bm{M}^{(t)}\cdot\bm{v}_{x'},
	\end{equation}
	Considering all the predicate matrices at each step and relational paths of different lengths, the final score \paulw{can} be rewritten with soft attention as below:
	\begin{align}
	\kappa(\bm S_\psi, \bm S_\varphi) &= \sum_{t'=1}^{T}s^{(t')}_\psi\left(\prod_{t=1}^{t'}\sum_{k=1}^\paulw{N} s^{(t)}_{\varphi,k}\bm{M}_k\right), \label{5}\\
	{\rm score(x,x')} &=
	{\bm{v}_{x}}^{\intercal}\kappa(\bm S_\psi, \bm S_\varphi)\bm{v}_{x'}, \label{eq:score}
	\end{align}
	where $T$ is the maximum reasoning steps, $\bm S_\psi = (s^{(t')}_\psi)_{t'}$, $\bm S_\varphi = (\bm s^{(t)}_{\varphi,k})_{t, k}$, term $s^{(t')}_\psi$ corresponds to attention weights \paulw{over} relational paths of length $t'$, and $s^{(t)}_{\varphi,k}$ to another attention weights on predicate matrix $\bm{M}_k$ used in the $t$-th step, and \paulw{$N$} denotes the total number of predefined predicates.

\subsection{Attention Module}\label{sec:attention module}
	In this section, we introduce the architecture of \paulw{the} attention \paulw{module}, a hierarchical stack of transformers, to generate the dynamic attention weights. 
	\paulw{Recall} a basic multi-head dot-product attention module (MHDPA) in \paulw{the transformer architecture} \citep{vaswani2017attention} \paulw{takes as inputs} the query, key and value representations: $\bm Q$, $\bm K$, $\bm V$. 
	MHDPA firstly computes the similarity or attention weights $\bm S$ between the query and the key, and then calculates the weighted value as output $\bm V'$\paulw{:
	\begin{align}
	& \textbf{\rm MHDPA}(\bm Q, \bm K, \bm V) = \bm S, \bm V' \\ 
	& \mbox{with } \bm S = {\rm softmax}\big(\frac{\bm Q \bm K^{\intercal}}{\sqrt{d}}\big) {\mbox{ and }} \bm V' = \bm S \bm V,
	\end{align}}%
	where $d$ is the dimension of $\bm K$. 
	
	We utilize this module to generate the attention weights $\bm S_\varphi$ and \paulw{$\bm S_\psi$}. 
	In fact, the symbolic states can be represented as a 3-dimensional tensor $\bm{M} \in [0,1]^{|\mathcal{X}| \times |\mathcal{X}| \times N}$, where $\mathcal X$ denotes the set of extracted objects and $N$ represents the numbers of predefined predicates. We transform tensor $\bm{M}$ into a matrix $\bm{M}_f \in [0,1]^{{|\mathcal{X}|}^2 \times N}$ at each time step. 
	Each row of matrix $\bm{M}_f$ represents a part of the symbolic state, which can be seen as an embedding of predicate. 
	In this way, the attention module can generate weights on predicates at different reasoning steps, taking consideration of the symbolic information of current RL state. 
	We firstly generate the query, key and value representation\paulw{s with multi-layer perceptrons} with
	$\bm{M}_f$ as \paulw{initial} input. 
	For convenience, we \paulw{define $\bm V^{(0)}_\varphi = \bm{M}_f$}. Then, we repeatedly use the output value from last step to generate the attention weights.  \paulw{The predicate attention submodule can be summarized as follows:
	\begin{align}
	& \bm Q^{(t)}_\varphi, \bm K^{(t)}_\varphi, \bm V^{(t)}_\varphi = {\rm FeedForward}_t(\bm V^{(t-1)}_\varphi), \\
	& \bm s^{(t)}_\varphi, \bm V^{(t+1)}_\varphi = \textbf{\rm MHDPA}(\bm Q^{(t)}_\varphi, \bm K^{(t)}_\varphi, \bm V^{(t)}_\varphi), 
	\end{align}
	where the superscript denotes the reasoning step.}
	Here, $s^{(t)}_\varphi$ represents the attention weights \paulw{over} predicates in the $t$-th hop reasoning and $\rm FeedForward$ means multi-perctron layer.
	For \paulw{the} path attention submodule, we reuse the output value of each time step in \paulw{the} predicate attention submodule. 
	During the iterative processing, the output value at each step embeds the information of paths of different lengths. 
	We simply use another transformer to generate the path attention weights $\bm S_\psi$.
	Let $\bm V_\phi={[\bm V^{(0)}_\varphi, \bm V^{(1)}_\varphi, \cdots, \bm V^{(t)}_\varphi]}^{\intercal}$.
	\begin{align}
	\bm Q_\psi, \bm K_\psi, \bm V_\psi &= {\rm FeedForward}(\bm V_\phi), \\
	\bm S_\psi, \paulw{\bm V'_\psi} &= \textbf{\rm MHDPA}(\bm Q_\psi, \bm K_\psi, \bm V_\psi),
	\end{align}
	
	\subsection{Policy Module}\label{sec:policy module}
	In this section, we build a policy module for generating DRL policies.
	\zhma{We denote an object set and a predicate set by $\mathcal{X}$ and $\mathcal{P}$ respectively. A predicate set is composed of both action predicates representing the fact of changing states, denoted by $\mathcal{P}_a$, and state predicates representing the fact of states, denoted by $\mathcal{P}_s$.}
	Let $x$ and $x'$ represent two entities from $\mathcal{X}$. We denote an action atom by $Act_a(x,x')$ where $Act_a$ is an action predicate from $\mathcal{P}_a$.
    Assuming there exists an oracle capable of extracting symbolic states at each time step, we can represent state and action with FOL. 
	Since a policy is a mapping from states to actions, \paulw{a} predicate at the last hop needs to be constrained to be an action predicate. 
	For every action predicate $Act_a$, we introduce a \paul{multi-layer perceptron} $\rm MLP_a$ to the output of \paul{the} reasoning module to induce the state-action value of action atom $Act_a(x,x')$:
	\begin{equation}
	Q(S, Act_a(x,x')) =
	({\bm{v}_{x}}^{\intercal} {\rm MLP}_a(\kappa(\bm S_\psi, \bm S_\varphi))\bm{v}_{x'}),
	\end{equation}
    \paulw{In} the context of FOL, we only consider \paulw{finite} action space. 
    To learn a deterministic policy, we update the policy module by minimizing the loss function described below where $D$ is a \paulw{replay buffer}.
	\begin{equation*}
	L(\theta)=\mathbb{E}_{(s, \hat a, r, s') \sim D}\left[\big(r + \gamma \, \paul{\max}_{\hat a'} Q(s', \hat a'; \theta) - Q(s, \hat a;\theta)\big)^\paulw{2}\right],
	\end{equation*}
	\paulw{where $\hat a$ and $\hat a'$ denote action atoms.}
	To learn a \paulw{stochastic} policy, we \paulw{use} the $softmax$ function on the state-action values and obtain probabilities over taking action atoms. Then, we can train {\ours} with any deep RL algorithms such as DQN \citep{Mnihet.al2015} or policy gradient methods (e.g., REINFORCE \citep{Williams1992simple}, PPO \citep{schulman2017proximal}).

\section{Experiments}

    \begin{figure*}[htbp]
    \centering
    \hspace{-2mm}
    \subfigure[Sample Play]{
    \centering
    \includegraphics[scale=0.27]{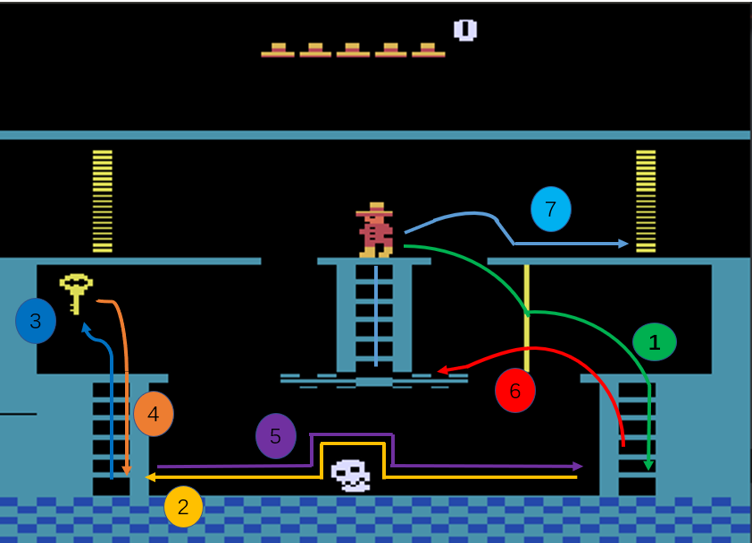}
    \label{optimal policy}
    }%
    \hspace{1mm}
    \subfigure[Learning Curve]{
    \centering
    \includegraphics[scale=0.27]{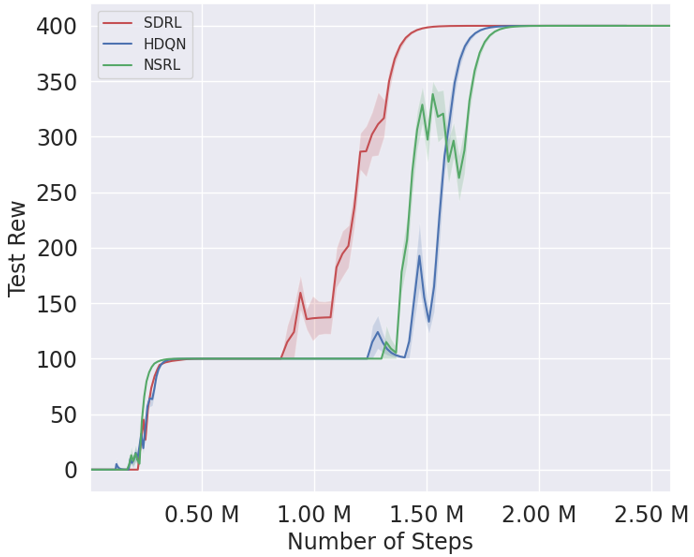}
    \label{learning curve}}
    \hspace{1mm}
    \subfigure[Blocks World]{
    \centering
    \includegraphics[scale=0.27]{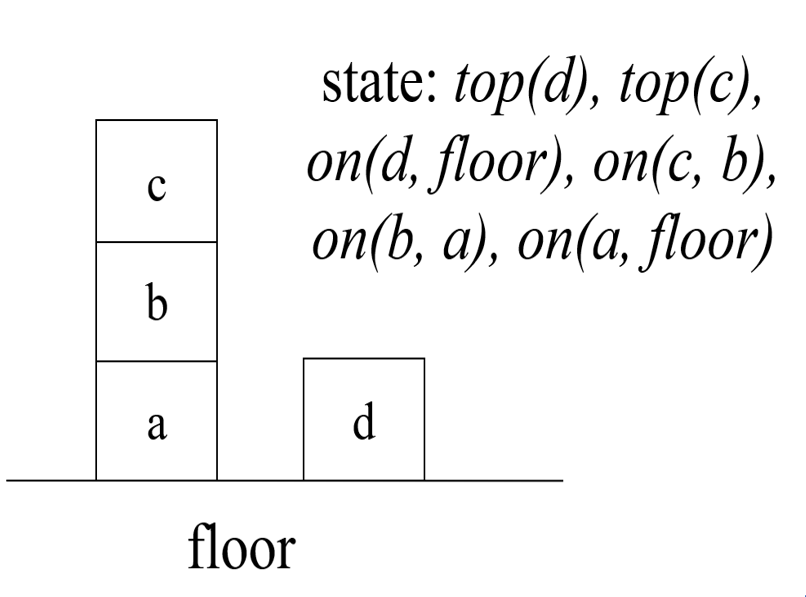}
    \label{block world}
    }
    \caption{\textbf{(a)} describes the optimal policy learnt by the agents. \textbf{(b)} illustrates the evaluating phase of SDRL, HDQN and NSRL. \textbf{(c)} depicts a sample of the block world domain with the state represented by FOL.}
    \end{figure*}

    In this section, \hankz{we evaluate our approach on two domains, i.e., Montezuma's Revenge and Block\paul{s} World Manipulation, in terms of expected returns, generalization ability, and interpretability.} 
    We measure the performance of our approach \paulw{in terms of} expected returns. 
    \paulw{H}igher returns indicate better performance. 
    To validate the generalization ability leveraged by symbol logic, we compute the expected returns the agents receive in unseen environments. 
    We \paulw{qualify a method as interpretable} if it can present the logical rules learned in the training process.  
    \paulw{We compare our proposition with relevant state-of-the-art algorithms in both domains.} \zhma{We set the learning rate to be 1e-4, \paulw{the} maximum reasoning steps to be 4 and the number of layers and heads in the attention module to be 2 and 4 separately.}
    \paulw{We describe each domain, the evaluation protocol, and the results next.}
    
    \subsection{Montezuma's Revenge}\label{description about Montezuma's Revenge}
	\hankz{We first evaluate our approach on \emph{Montezuma’s Revenge}, an ATARI game with sparse, delayed rewards, which is also used by \citet{kulkarni2016hierarchical}. 
	\paulw{In this game, t}he player \paulw{navigates} through several rooms \paulw{in order to collect} treasures. 
	We conduct our experiment based on the first room shown in \paulw{Figure}~\ref{optimal policy}. 
	In this room, the player needs to first fetch a key\paulw{, then} navigate to the \paulw{right} door and pass through it. 
	If the player successfully fetches the key\paulw{, s}he receive\paulw{s} a reward (+100). 
	If \paulw{she} successfully navigates to the door and pass\paulw{es} through it, \paulw{s}he receive\paulw{s} another reward (+300).}

    \subsubsection{Symbolic Representation}
    The symbolic domain knowledge we use is based on 6 pre-defined locations: 
    \textbf{\textit{middle ladder, \paulw{(right)} door, left of rotating skulls, lower left ladder, lower right ladder}} and \textbf{\textit{key}}.
    One mobile object, the \textbf{\textit{man}} in red, is also introduced. 
    We introduce 6 predicates: \textbf{AtSpot, WithObject, \paulw{WithoutObject}, PathExist, KeyToDoor} and an action predicate \textbf{Move}. 
    Atom \textbf{AtSpot($x$, $y$)} means \paulw{object} $x$ is currently at location $y$. 
    \textbf{WithObject($x$, $y$)} means object $x$ possesses object $y$ and \textbf{\paulw{WithoutObject}($x$, $y$)} is the opposite. 
    \textbf{PathExist($x$, $y$)} means \paul{a} path from location $x$ to location $y$ exists and \textbf{KeyToDoor(\textit{key}, \textit{door})} means possessing a key is the \paulw{precondition} to open a door. 
    \textbf{Move($x$, $y$)} means move object $x$ to location $y$. 
    To represent a state with the symbolic predicates and objects defined above, we assume there is a pre-trained oracle capable of answering queries whether a specific atom is true given high-dimensional images as input.

    \subsubsection{Setup}
    We compare our approach with HDQN \citep{kulkarni2016hierarchical} and SDRL \citep{Daoming.el2018} as baselines. 
    We implement \paulw{our architecture} {\ours} and HDQN with an option-based hierarchical reinforcement learning framework \paulw{similar to} SDRL. This framework is split into two levels, meta controller (high level) and \paulw{action} controller (low level). The meta controller assigns a task to be achieved by the \paulw{action} controller. 
    The only difference \paulw{between} these agents is the way to induce \paulw{a} policy in the high level. 
    SDRL requires a symbolic transition model (expert knowledge) and a planner to induce an option trace. 
    HDQN utilizes an end-to-end neural network to induce the higher level policy while {\ours} performs \paulw{neuro-symbolic} reasoning. 
    In terms of the low level, all the agents reuse the controller architecture in \citet{kulkarni2016hierarchical} and we set the maximum interaction length to be 500. \zhma{To facilitate the learning process, we define the reward function below for training and use the original reward setting described in Section \ref{description about Montezuma's Revenge} for testing.} The controller receives \paulw{a} reward \paulw{of} -0.1 at every step and +10 when achieving the assigned goal. If the controller fails the game or lose its life, it will receive another reward -5. The meta controller will receive -0.5 reward after each decision.
    We jointly train the two levels of these algorithms with \paulw{the Deep} Double Q-Learning algorithm \citep{vanhasselt2016deep} and prioritized replay buffer \citep{schaul2015prioritized}. 

    \subsubsection{Results}
    
    We present the optimal policy learned as shown in \paulw{Figure}~\ref{optimal policy}. These agents sequentially learn \paulw{Tasks $1$ to $3$} to get +100 reward and explore other \paulw{Tasks $4$ to $7$} and finally \paulw{converge} to +400 reward. 
    We \paulw{estimate} expected returns from 8 runs and present the results in \paulw{Figure}~\ref{learning curve}. 
    The performance of these approaches are similar. 
    It takes nearly 1.5M steps for SDRL to \paulw{converge} to the optimal performance (+400 reward) while 1.8M steps \paulw{are needed} for {\ours} and HDQN. 
    Due to the use of \paulw{the $\epsilon$-greedy exploration strategy} in \paulw{Deep Double}  Q-learning, both {\ours} and HDQN take another 0.3M steps on exploration than the planner-based method SDRL.
    The use of a symbolic planner \paulw{with the formalization of the planning problem guides the learning agent} to induce an increasing\paulw{ly better} plan, \paulw{explaining} the fast\paulw{er converge}nce of SDRL.
    However, {\ours} still performs competitively and similarly \paulw{to} the model-free method HDQN. Both of them start to explore \paulw{how} to fetch a key (+100 reward) and open a door (+400 reward) nearly at the same time.
    \begin{table}[t]
        \centering
        \small
        \caption{Comparison of the three agents in terms of expert knowledge dependency}
        \begin{tabular}{lc}
            \toprule
             Method &  Expert Knowledge\\
             \midrule  
             {\ours}  &   6 locations, 6 predicates\\
             \midrule  
             SDRL  &   6 locations, 5 predicates, 1 transition model\\
             \midrule  
             HDQN  &   ---\\
            \bottomrule
        \end{tabular}
        \label{expert knowledge dependency}
    \end{table}
    \paulw{By design, SDRL can provide an interpretable plan. 
    However, we argue that this approach does not scale, since SDRL requires a full description of the planning problem.}
    To compare the \paulw{dependency on} expert knowledge, we enumerate in Table~\ref{expert knowledge dependency} the symbolic knowledge used in each method. 
    \paulw{Obviously} HDQN does \paulw{not} use any domain knowledge and thus \paulw{prevents any} interpretability with logical rules. 
    Although {\ours} \paulw{uses} one \paulw{extra} predicate \paulw{compared to} SDRL, {\ours} \paulw{depends less on} expert knowledge than SDRL since design\paulw{ing} a symbolic transition model in complicated environments \paulw{is much harder} than predicates. 
    Therefore, it \paulw{is} easier for {\ours} to scale \paulw{to more complex problems} than SDRL. 
    We leave the discussion about \paulw{the} interpretability of {\ours} \paulw{to} Section \ref{interpretable section}.

    \subsection{World Manipulation}
    We validate the generalization ability of {\ours} in the Block\paul{s} Manipulation Environment \paulw{used by \cite{jiang2019neural}}. In this environment, the agent is required to finish three tasks: \textbf{\textit{STACK}}, \textbf{\textit{UNSTACK}} and \textbf{\textit{ON}}. In the \textbf{\textit{STACK}} task, \paulw{the} blocks need to be stacked into a column while \paulw{they need to be put} on the floor in the \textbf{\textit{UNSTACK}} task. In the \textbf{\textit{ON}} task, a specific block is required to \paulw{be} put on another one. In all the tasks, the agents is trained with only 4 blocks while tested with 5 or even more blocks. Three predicates \paulw{(i.e., \textbf{On}, \textbf{Top}, and an action predicate \textbf{Move})} are \paulw{used} to represent the symbolic states and actions. Another predicate \textbf{GoalOn} is also introduced in task \textbf{\textit{ON}} to specif\paulw{y} the goal\paulw{, i.e.,}  \textbf{{GoalOn(a,b)}} \paulw{means that} block $a$ \paulw{should be moved} on entity $b$\paulw{, which can be a block or the floor}. \paulw{Figure}~\ref{block world} shows the state \textit{((a,b,c), (d))} and its symbolic representation.

\begin{table*}[!h]
    \centering
    \caption{\paulw{Expected returns of different agents in training/test environments. The first row provides each agent's category.
    The first 2 columns list the tasks and their instances.
    The next 5 show the performance of the agents, in addition to} the optimal returns computed by value iteration \paulw{(VI).}}
    \begin{tabular}{lllllll}
    \toprule
    Type    &         & Rules Learning       & Rules Given    & No Rules &  No Rules &                \\
    \midrule 
	Method	&             & {\ours}                 & NLRL           & NLM      & MLP   & VI        \\
    \midrule 
	UNSTACK & training    & \bm{$0.939 \pm 0.004$} & $0.935 \pm 0.011$ & $-0.773 \pm 0.495$ & $0.934 \pm 0.012$ & 0.940\\
	
	& swap top 2          & \bm{$0.939 \pm 0.005$} & $0.935 \pm 0.010$ & $-0.777 \pm 0.492$ & $0.920 \pm 0.025$ & 0.940\\
	
	& 2 columns      & \bm{$0.960 \pm 0.000$}  & $0.956 \pm 0.009$ & $-0.424 \pm 0.749$ & $-0.951 \pm 0.203$ 
	& 0.960\\
	
	& 5 blocks       & \bm{$0.919 \pm 0.005$} & $0.910 \pm 0.016$ & $-0.953 \pm 0.183$ & $0.900 \pm 0.256$
	& 0.920\\
	
	& 6 blocks       & \bm{$0.894 \pm 0.017$} & $0.884 \pm 0.020$ & $-0.979 \pm 0.044 $ & $0.862 \pm 0.033$
	& 0.900\\
	
	& 7 blocks       & \bm{$0.864 \pm 0.020$} & $0.855 \pm 0.026$ & $-0.980 \pm 0.000 $ & $0.762 \pm 0.098$
	& 0.880\\
    
    \midrule  
			
	STACK   & training       & \bm{$0.940 \pm 0.003$} & $0.889 \pm 0.046$ & $0.129 \pm 0.702$ & $0.937 \pm 0.009$ & 0.940\\
	
	& swap right 2   & \bm{$0.940 \pm 0.004$} & $0.889 \pm 0.045$ & $0.156\ \pm 0.688$ & $0.937 \pm 0.010$
	&0.940\\
	
	& 2 columns      & \bm{$0.939 \pm 0.028$} & $0.919 \pm 0.055$ & $0.182 \pm 0.709$ & $-0.980 \pm 0.000$
	&0.940\\
	
	& 5 blocks       & \bm{$0.917 \pm 0.017$} & $0.863 \pm 0.053$ & $-0.437 \pm 0.699$ & $-0.980 \pm 0.000$
	&0.920\\
	
	& 6 blocks       & \bm{$0.878 \pm 0.139$} & $0.834 \pm 0.069$ & $-0.772 \pm 0.491$ & $-0.980 \pm 0.000$
	&0.900\\
	
	& 7 blocks       & \bm{$0.826 \pm 0.210$} & $0.791 \pm 0.134$ & $-0.912 \pm 0.286$ & $-0.923 \pm 0.257$
	&0.880\\

    \midrule  

	ON      & training       & \bm{$0.917 \pm 0.008$} & $0.913 \pm 0.012$ & $-0.823 \pm 0.432$ & $0.512 \pm 0.468$ & 0.920\\
	
	& swap top 2     & $0.907 \pm 0.018$ & \bm{$0.915 \pm 0.010$} & $-0.817 \pm 0.437$ & $0.840 \pm 0.067$
	& 0.920\\
	
	& swap mid 2     & \bm{$0.916 \pm 0.009$}  & $0.915 \pm 0.011$ & $-0.859 \pm 0.383$ & $0.663 \pm 0.239$
	& 0.920\\
	
	& 5 blocks       & \bm{$0.888 \pm 0.018$} & \bm{$0.888 \pm 0.018$} & $-0.939 \pm 0.231$ & $-0.910 \pm 0.303$
	& 0.900\\
	
	& 6 blocks       & $0.852 \pm 0.030$ & \bm{$0.866 \pm 0.020$} & $-0.977 \pm 0.063$ & $-0.980 \pm 0.000$
	& 0.880\\
	
	& 7 blocks       & $0.798 \pm 0.048$ & \bm{$0.839 \pm 0.019$} & $-0.980 \pm 0.000$ & $-0.980 \pm 0.000$
	& 0.860\\

    \bottomrule
    \end{tabular}
    \label{result of block world}
    \end{table*}

    \subsubsection{Setup}
    The settings of \paulw{the} environments are the same as \paulw{in \citep{jiang2019neural}}. 
    In total, there are \paulw{a maximum of} 7 blocks labeled as \textit{(a, b, c, d, e, f, g)} and one entity labeled as \textit{floor}. 
    \paulw{The agent} is asked to \paulw{operate} on these entities to finish the tasks. 
    In the interaction process, the agent receives a reward of -0.02 at every step and gets a reward of 1 after finishing its task and the maximum length of interaction is set to be 50. 
    If the action is invalid like \paulw{\textbf{Move}$(floor, a)$}, the state will not be changed.
    In order to test the generalization ability, the agent is trained in environments with 4 blocks while tested in environments with more than 4 blocks. 
    In \paulw{the} \textbf{\textit{UNSTACK}} task, the agent is trained with a single column of blocks like \textit{((a, b, c, d))}. 
    We swap the top 2 blocks or divide the blocks into 2 columns for testing. 
    For \paulw{the} \textbf{\textit{STACK}} task, the initial state is like \textit{((a),(b),(c),(d))} in \paulw{the} training environment while \textit{((a), (b), (d), (c))} and \textit{((a, b), (d, c))} \paulw{are used} in generalization tests. 
    For \paulw{the} \textbf{\textit{ON}} task, the initial state in \paulw{the} training environment is \textit{((a, b, c, d))} and the goal is to put block $a$ on block $b$. 
    We also swap the top 2 blocks or middle 2 blocks for testing. 
    \paulw{Besides, w}e randomly choose 4 blocks from the total 7 blocks to replace the above mentioned blocks in \paulw{the} training environments and produce more training cases.
    In all the tasks, the agent is required to test in unseen environment with $5\sim7$ blocks. 
    The test environments with over 4 blocks are the same in \paulw{the} \textbf{\textit{UNSTACK}} and \textbf{\textit{ON}} task\paulw{s}, which \paulw{are} $((a, b, c, d, e))$, $((a, b, c, d, e, f))$, and $((a, b, c, d, e, f, g))$. 
    The initial state\paulw{s} in test environments with more than 4 blocks for \paulw{the} \textbf{\textit{STACK}} task are \textit{((a),(b),(c),(d),(e)), ((a),(b),(c),(d),(e),(f))} and \textit{((a),(b),(c),(d),(e),(f),(g))}.
    We compare {\ours} with NLRL \citep{jiang2019neural}, NLM \citep{Dong.el2019}, and \paulw{a Multi-Layer Perceptron} (MLP) in these tasks. 
    NLRL requires rule templates to generate possible rules but NLM and MLP \paulw{do not allow the extraction of} rules after learning. For this reason, we classify these algorithms into three types: Rules Learning, Rules Given and No Rules.
    The MLP agent has 2 hidden layers with 20 units using RELU \citep{nair2010rectified} activation functions. 
    \zhma{
    \paulw{Following NLRL, we train MLP, and {\ours} with the PPO algorithm \citep{schulman2017proximal} and use} generalized advantages ($\lambda=0.95$) \citep{schulman2015high}.
    \paulw{For the three architectures, we use the same critic network consisting of one 20-unit hidden layer.}
    }

    \begin{table}[t]
	    \caption{\hankz{\paulw{Comparison of} the four agents in terms of inference time and memory cost.}}
        \begin{tabular}{ccccc}
            \toprule
             Method & {\ours} &  NLRL & NLM & MLP\\
             \midrule  
             Inference Time (h) & $1.219$  & $5.421$ & $1.396$ & \bm{$0.168$}\\
             \midrule  
             Memory Cost (GB)  & $2.715$ & $7.340$ & \bm{$2.056$}  & $2.682$ \\
            \bottomrule
        \end{tabular}
        \label{comparision of scalability}
    \end{table}

	\subsubsection{Results}
	\hankz{We exhibit the \paul{averages and standard deviations over} 1000 repeated evaluations of \paulw{the} three blocks world manipulation tasks in Table~\ref{result of block world}. We employ a stochastic policy for all the agents in the evaluation phase.}
	\zhma{From Table~\ref{result of block world}, we can see that the MLP agent achieves near-optimal performance in the training environments of \textbf{\textit{UNSTACK}} and \textbf{\textit{STACK}} task. However, without any design of logic reasoning in more difficult and complex tasks, the MLP agent fails in most of the testing environments of \textbf{\textit{STACK}} and \textbf{\textit{ON}} task. Although NLM introduces a logical architecture inductive bias, it performs worse than MLP. In general, NLRL outperforms NLM and MLP in terms of expected returns in the three tasks. }
	The logical rules of higher confidence learned from a given rule set in \paulw{the} training environment can be directly reused in \paulw{the} test environments, \paulw{contributing to its} great generalization ability in unseen environments.
	Besides, {\ours} achieves competitive performance compared to NLRL not only in training but also in test environments. 
	\zhma{In task \textbf{\textit{UNSTACK}}, {\ours} achieves near-optimal performance, about 0.05 returns higher than NLRL in training environments and 0.1 higher in test environments. {\ours} performs more stably than NLRL. The standard deviations of {\ours} is about 0.005 less than NLRL.
	In the difficult task \textbf{\textit{STACK}}, {\ours} achieves the optimal returns in training environment. The performances of both {\ours} and NLRL decrease gradually in environments with increasing number of blocks. However, {\ours} can still achieve 0.917, 0.878 and 0.826 average returns in environments with 5, 6, and 7 blocks separately, about 0.03 higher than NLRL. 
	In the first four environments of task \textbf{\textit{ON}}, {\ours} performs closely to NLRL with the gap between the returns being less than 0.008 and the standard deviations less than 0.005. However in the remaining tests, NLRL performs better.}
    
    To compare the scalability between these agents, we \paul{evaluate the computation inference time and memory cost for each task averaged} over 1000 tests.
	Although the MLP agent uses least inference time, it performs badly in the test environments and so as the NLM agent.
	Table \ref{comparision of scalability} illustrates \paulw{that {\ours} scales better than NLRL. Indeed, NLRL takes almost 2.7 times more memory and 4.5 times more inference time than {\ours}. The high memory requirement of NLRL, increasing exponentially with the number of predicates, prevents it to scale to a complex domain like Montezuma's Revenge. These results illustrate that {\ours}, without human-designed rules, can also achieve competitive performance with improved scalability.}
	Athough NLM uses \paulw{a} symbolic representation, it can not extract logical rules and \paulw{neither can} the MLP agent. Therefore, the polic\paulw{ies} learned by these two agents are \paulw{lack} of interpretability. 
	The logical rules of higher confidence in NLRL are generated from human designed rules templates and thus \paulw{are} highly interpretable. 
	We leave the discussion about \paulw{the} interpretability of {\ours} in \paulw{the} Block\paul{s} World domain \paulw{to the} next section.
    
    \begin{table*}[t]
        \centering
        \caption{Logical Rules extracted from {\ours}\paulw{.}}
        \begin{tabular}{ll}
            \toprule
             Domain &  Logical rules\\
             \midrule  
             Block\paul{s} World &  1. \textit{$Move(X,Y) \leftarrow GoalOn(X,Y)$} \\
                         &  2. \textit{$Move(X,Z) \leftarrow Top(X,X) \land On(X,Y) \land On(Y,Z)$} \\
                         &  3. \textit{$Move(X,M) \leftarrow On(X,Y) \land On(Y,Z) \land On(Z,M)$} \\
                         &  4. \textit{$Move(X,Y) \leftarrow Top(X,X) \land GoalOn(X,Y) \land Top(Y,Y) $} \\
                         &  5. \textit{$Move(X,Z) \leftarrow On(X,Y) \land GoalOn(Y,X) \land On(X,Y) \land On(Y,Z)$} \\
             \midrule  
             Montezuma's Revenge  &  6. \textit{$Move(man,key) \leftarrow \paulw{WithoutObject}(man,key) $} \\
                         &  7.  \textit{$Move(man,door) \leftarrow WithObject(man,key) \land KeyToDoor(key,door)$} \\
            \bottomrule
        \end{tabular}
        \label{logical rules}
    \end{table*}

    \subsection{Interpretable Policy}\label{interpretable section}
    In this section, we present the logical rules learned by {\ours} in the domain of Montezuma's Revenge and Block\paul{s} World Manipulation. We visualize the attention weights for every predicates at each reasoning step and for path\paulw{s} of different length\paulw{s}. Then, we multiply these predicate attention weights sequentially. The \paulw{results of the product are multiplied by the attention weights of the path of the} corresponding length. 
    We \paulw{interpret these} results as the confidence of \paulw{the} corresponding logical rules. 
    Since {\ours} can only learn chain-like rules, we \paulw{manually} select some of the chain-like logical rules of highest attention weights as shown in Table~\ref{logical rules}. 
    For example, in \paulw{the} \textbf{\textit{ON}} task, \paulw{if the current state satisfies rules 1 or 4, it is} most likely to take action \textbf{\textit{Move(X,Y)}}.
    \paulw{In the Atari Game, a likely reason to choose to get the key as a task is when the agent does not have it}. 
    \paulw{The} predicate \textbf{KeyToDoor} \paulw{that we introduced} to embed the knowledge that a key is important to open a door \paulw{also improves} the interpretability of \paulw{the learned rules, i.e.,} \paulw{rule 7, which chooses a door as a task}. 
    These \paulw{extracted} logical rules are not always true and \paulw{need to be} selected by \paulw{a} human but \paulw{they do provide a certain interpretation for why an action is chosen}. 
    \paulw{In any case, {\ours} is arguably more interpretable than MLP and NLM}. {\ours} provides \paulw{a novel} way to generate chain-like rules without human designed templates, improving flexibility and saving human labour.

\section{Conclusion}	
	In this paper, we propose a \paul{novel} framework performing \paul{neural-logic} reasoning to enable interpretability by visualizing the relational paths to tasks. 
	\paulw{Exploiting multi-hop reasoning, attention mechanism, and hierarchical reinforcement learning, our approach can solve large-sized complex problems like Montezuma's Revenge, in contrast to other recent neuro-symbolic approaches.}
	\paul{Compared to other black-box} methods, \paulw{our approach naturally operates with} symbolic knowledge while achieving comparable performance and preserving \paul{interpretability}.
	\paulw{As future work, our framework can be extended to allow more expressive rules such as tree-like or junction-like rules \citep{yuan2019}. Such extensions could improve further the performance and the interpretability of {\ours}.
	Another interesting and important research direction is to learn the predicates directly from high-dimensional inputs (e.g., images).}

\begin{contributions} 
    Briefly list author contributions.
    This is a nice way of making clear who did what and to give proper credit.

    H.~Q.~Bovik conceived the idea and wrote the paper.
    Coauthor One created the code.
    Coauthor Two created the figures.
\end{contributions}

\begin{acknowledgements} 
    Briefly acknowledge people and organizations here.

    \emph{All} acknowledgements go in this section.
\end{acknowledgements}


\end{document}